\documentclass[runningheads]{llncs}
\usepackage{amsmath,epsfig}
\usepackage{xcolor}
\usepackage{multirow}
\usepackage[T1]{fontenc}
\usepackage{graphicx}
\usepackage{algorithm}
\usepackage{algorithmic}

\begin{document}
\title{Learning to Synthesize Graphics Programs for Geometric Artworks}
%
%
\author{Qi Bing\inst{1}\orcidID{0009-0006-4032-9873} \and
Chaoyi Zhang\inst{1}\orcidID{0000-0001-8492-9711} \and
Weidong Cai\inst{1}\orcidID{0000-0003-3706-8896}}
\authorrunning{Q. Bing et al.}
%
\institute{School of Computer Science, The University of Sydney, NSW 2006, Australia}
%
\maketitle              
\begin{abstract}
Creating and understanding art has long been a hallmark of human ability. When presented with finished digital artwork, professional graphic artists can intuitively deconstruct and replicate it using various drawing tools, such as the line tool, paint bucket, and layer features, including opacity and blending modes. While most recent research in this field has focused on art generation, proposing a range of methods, these often rely on the concept of artwork being represented as a final image. To bridge the gap between pixel-level results and the actual drawing process, we present an approach that treats a set of drawing tools as executable programs. This method predicts a sequence of steps to achieve the final image, allowing for understandable and resolution-independent reproductions under the usage of a set of drawing commands. Our experiments demonstrate that our program synthesizer, Art2Prog, can comprehensively understand complex input images and reproduce them using high-quality executable programs. The experimental results evidence the potential of machines to grasp higher-level information from images and generate compact program-level descriptions.

\end{abstract}
\section{Introduction}
\label{sec:intro}

\begin{figure}[t]
\begin{minipage}[b]{0.55\linewidth}
  \centering
  \centerline{\epsfig{figure=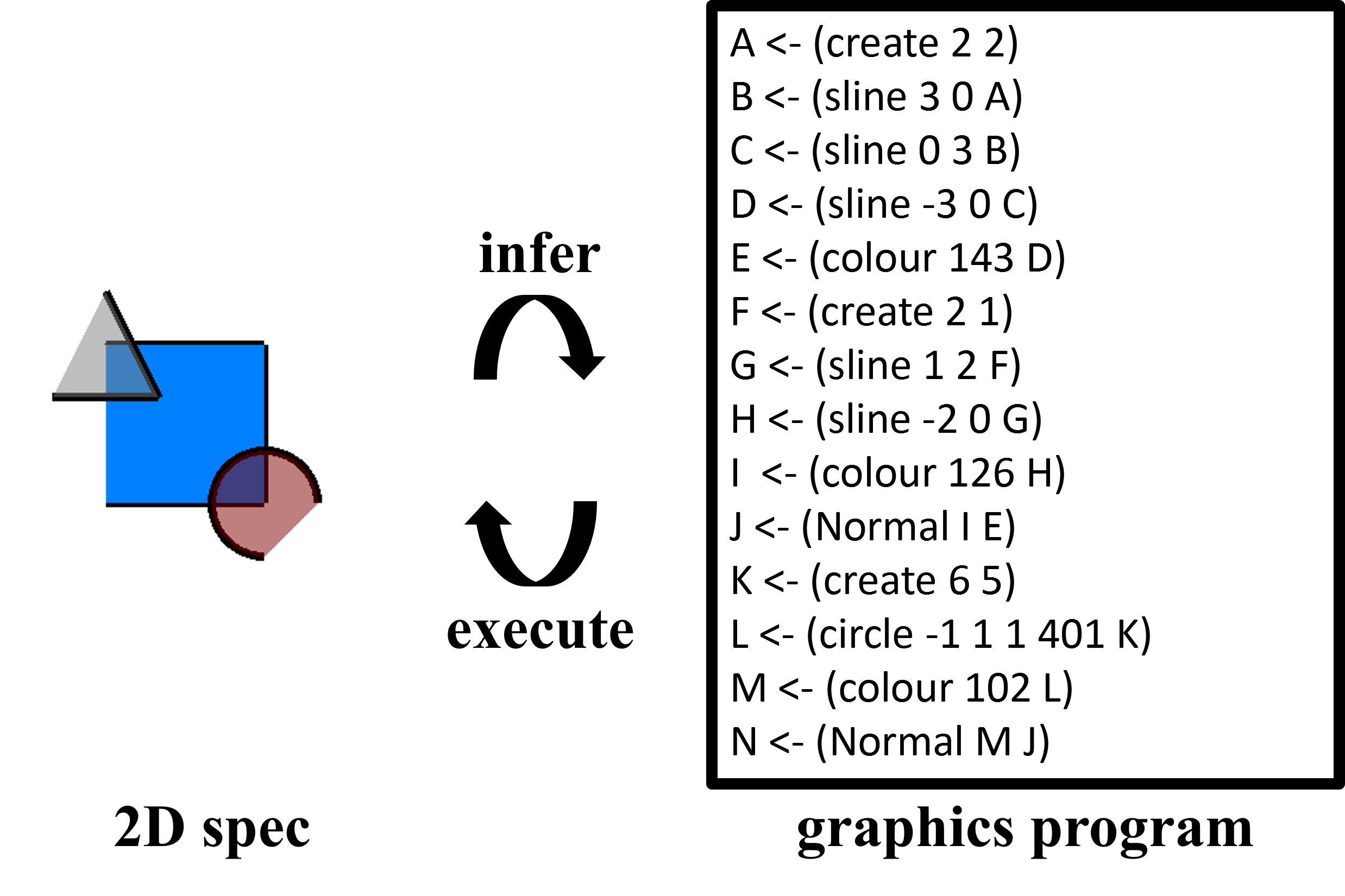,width=6.5cm}}
  \centerline{(a)}\medskip
\end{minipage}
\begin{minipage}[b]{0.44\linewidth}
  \centering
  \centerline{\epsfig{figure=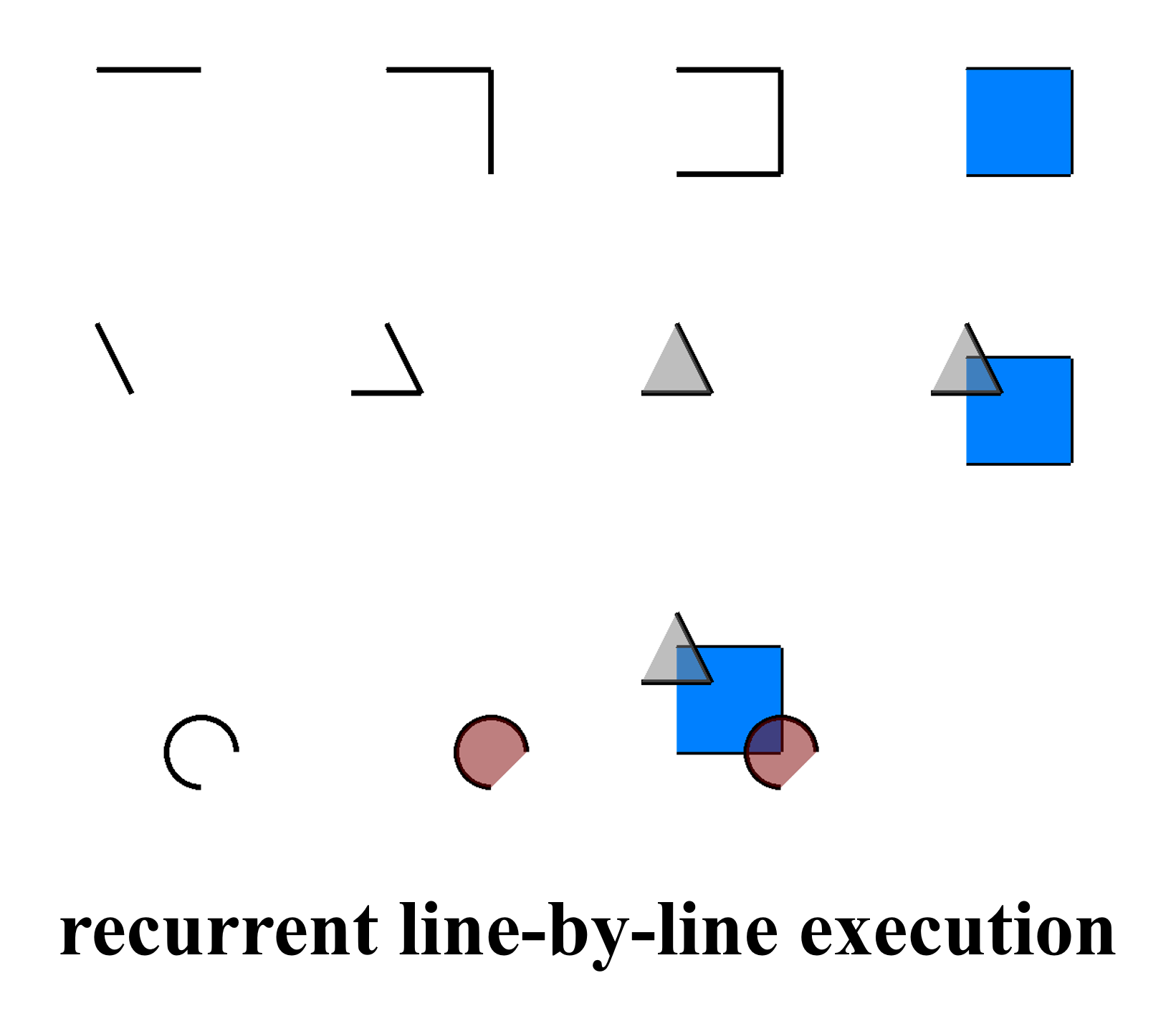,width=5.0cm}}
  \centerline{(b)}\medskip
\end{minipage}
\caption{(a): A complex 2D graphic composed of lines and colors can be represented and reproduced by an executable program. (b): The program comprises individual lines of code, each corresponding to a specific drawing command. Executing these commands sequentially reveals the process of constructing the graphic, closely mirroring an artist's workflow.}
\label{fig:res}
\end{figure}

Humans can easily understand the procedure that generates an image, no matter the drawing or characters, which necessitates understanding its underlying structure. However, inferring the drawing process from only the final image presents significant challenges. These challenges stem primarily from the occlusion of shapes and inherent ambiguity, as multiple interpretations can often be equally valid. Recent research has proposed various definitions for the process leading to the final image, including sketch colorization~\cite{tseng2020artediting,sketch_to_art,Filling2021zhang,xiang2022adversarial}, color segmentation~\cite{Akimoto_2020_CVPR} and time-lapse video generation~\cite{time_lapse}. 
Though these methods accomplished their tasks, their results still suffer from low resolution, distortion and noise to different degrees. 
There are also similar methods that aim to bridge the gap between images and other forms of description, such as vector graphics, by utilizing different types of parametric primitives, such as closed paths~\cite{Im2Vec,LIVE} or strokes~\cite{PaintTF,StylizedNeural}. Though these works produce promising vector-based results, they do not target to reason their generations. Instead, they approach the process more akin to vector-level segmentation.
To address these issues, we design a graphics program that can comprehensively represent the drawing process, mirroring the methods used by artists with digital drawing tools (e.g., straight or curved lines, paint buckets and layer blendings). By introducing an executable graphics program, images can be represented as structured drawing commands, enabling their reconstruction at any resolution (see Fig.~\ref{fig:res} as an example). Beyond its reconstruction capabilities, our proposed graphics program offers a representation of graphics that is not only readable and editable but also semantically meaningful. This makes it an ideal candidate for further applications, including drawing instruction.

Our work builds upon recent developments in graphics program synthesis~\cite{NEURIPS2019_50d2d226,Sharma_2018_CVPR,Feser2023InductivePS}, which have demonstrated the potential of program synthesis in decomposing complex shapes into a series of commands. However, the recurrent inference of the drawing process for colored images remains largely unexplored, and is the main focus of this paper.
Unlike most vectorization-based methods~\cite{LIVE,DiffVG,PaintTF,StylizedNeural,Im2Vec}, our approach does not rely on a differentiable rasterizer for path optimization or supervision in the pixel space. Instead, we train a program synthesizer to generate codes directly from a single image input. Our experimental results indicate that Art2Prog outperforms state-of-the-art optimization-based vectorization approaches in reconstruction accuracy while executing the inferred programs. Additionally, our method is capable of representing more complex graphics compared to existing graphics program synthesizers.

In summary, the contributions of this paper are threefold:
\begin{itemize}
  \item 
  As shown in Fig.~\ref{fig:res}, we define a colored graphics program that emulates an artist's workflow. This program is comprehensible to humans and capable of generating images at any resolution, thus efficiently bridging the bidirectional gap between programs and images.
  \item 
  We develop Art2Prog, a novel GPT-2~\cite{gpt2} based program synthesizer for generating colored graphics programs from a single input image. This architecture demonstrates the feasibility of capturing high-level information such as the number of layers, enclosed shapes, layer overlaps, and color blending modes, all through the inference of a complex 2D graphics program.
  \item 
  We evaluate our performance with state-of-the-art program synthesizers and optimization-based approaches in image vectorization. The experimental results show that Art2Prog outperforms existing works, achieving the highest reconstruction accuracy while also producing high quality program-level explanation.
\end{itemize}

\section{Related Works}
\subsection{Graphics program synthesis}
The task of learning to synthesize 2D graphics programs is not novel,  
with numerous recent papers being focused on reproducing 2D binary shapes. \cite{NEURIPS2019_50d2d226,Sharma_2018_CVPR,Feser2023InductivePS} primarily focused on reproducing CSG-based shapes, which are binary representations of solid shapes formed by applying boolean operations to simple shapes like circles and rectangles on the canvas. These methods successfully reconstructed solid shapes comprising up to 20 objects. Similarly, \cite{lego} constructed complex shapes by stacking a pre-defined binary shape (bricks of different lengths) built on the idea of Lego bricks. Written in a subset of \LaTeX, \cite{EllisRST18} defined programs as line shapes (e.g., circle, rectangle and straight line) rendered on an empty canvas. Building on this concept, \cite{dreamcoder} conceptualized the program as controlling a `pen' that draws binary lines on an empty canvas, and \cite{CAD_sketch} followed a similar approach for reconstructing CAD sketches by sequentially drawing lines. Additionally, \cite{spiral} introduced parameterized brushstrokes as programs and generated blurry paintings from photos.

To the best of our knowledge, our proposed Art2Prog is the  first work that explicitly targets the inference of complex 2D graphics programs that include lines, colored surfaces, and overlapping layers with different blending modes.

\subsection{Image vectorization}
Different from image rasterization, vectorization is inherently more complex due to the potential non-uniqueness of its results. Traditional methods~\cite{Bayesian2008,Ardeco2006,Cartoons2005,Codec2005,curvilinear2009,Hilaire2006RobustAA} normally build specific algorithm-based methods that rely on image segmentation to conduct vectorization. To address this issue, recent research has tried to leverage the power of learning-based approaches. Studies \cite{LIVE,DiffVG,PaintTF,StylizedNeural} approached vectorization by optimizing a fixed number of parametric strokes relying on differentiable rasterizers. However, these methods needed to fully account for shape semantics, leading to redundant and inaccurate vectorization. Meanwhile, \cite{Im2Vec} trained an encoder-decoder model without supervision from vector ground-truth. However, it still relied on a differentiable rasterizer for loss backpropagation. Similarly, \cite{SVG-VAE,DeepSVG} attempted vector reconstruction using Variational Autoencoders (VAEs) but failed to reach accurate results. \cite{ClipGen} implemented category-conditioned image vectorization through a two-module network, adding one layer of solid color at a time. Their recent works highlight the potential of deep architectures to capture higher-level structural information in image vectorization. 

Different from most existing vectorization approaches, Art2Prog does not rely on differentiable rasterization. Instead, we directly synthesize a program from the image input and execute it to accurately reconstruct the image. Furthermore, our method not only generates resolution-independent vector graphics but also comprehensively describes the image through human-understandable programs.

\begin{table}
\begin{center}
\caption{
The domain-specific language (DSL) for our 2D graphics. 
}
\vspace*{1mm}
\label{table:DSL}
\begin{tabular}{lll}
\hline
\noalign{\smallskip}
Program $P$ &$\rightarrow$ &\ \ $ O\ |\ O(O)\ |\ L(O,\ O)$ \\
Operation $O$  &$\rightarrow$  &\ \ Create($x=N,\ y=N$) \\
&& $|$ Straight($x=N,\ y=N,\ O$) \\
&& $|$ Circle($x=N,\ y=N,\ r=R,$\\
&& \qquad \ \ \ \ \ $dir=D,\ O$) \\
&& $|$ Fillcolor($color=C, O$) \\
Layer $L$  &$\rightarrow$& \ \ Normal($O1=O,\ O2=O$) \\
&& $|$ Multiply($O1=O,\ O2=O$)\\ 

Position $N$ &$\rightarrow$& \ \ integers within range of $[-8:8]$ \\
Radius $R$ &$\rightarrow$& \ \ integers within range of $[1:4]$ \\
Direction $D$ &$\rightarrow$& \ \ $True\ |\ False$ \\
color $C$ &$\rightarrow$& \ \ integers within range of $[1:54]$ \\
\hline
\end{tabular}
\end{center}
\end{table}

\section{Graphics Programs}
\label{sec:DSL}
This section defines the domain-specific language (DSL) used for our 2D graphics program. As depicted in Table~\ref{table:DSL}, our graphics program is structured hierarchically. A final image comprises several overlapping layers, each utilizing one of two distinct color blending modes. To construct each layer, multiple drawing commands must be executed sequentially:
\begin{enumerate}
    \item A $Create~(x,\ y)$ command that initiates a new layer on the current canvas and sets the starting position at coordinates $(x,\ y)$. 
    \item Multiple $Straight~(x,\ y,\ O)$ commands that draw continuous straight lines from the last position to a relative position $(x,\ y)$. These commands also help define enclosed shapes that can be filled with color.
    \item Multiple $Circle~(x,\ y,\ r,\ dir,\ O)$ commands that draw circular arcs. These arcs extend from the current position to a specified relative position $(x,\ y)$, defined by a radius $r$ and a direction $dir$, which can be either clockwise or counterclockwise.
    \item A $Fillcolor~(C,\ O)$ command that applies the color $C$ to an enclosed shape. This enclosed shape is determined by the line path resulting from an earlier operation $O$, which is defined by the arrangement of lines in the current layer. 
\end{enumerate}

$Normal~(O,\ O)$ or $Multiply~(O,\ O)$ employs two distinct layer blending modes to connect a pair of object layers, denoted as $O$.
All of the variables $N$, $R$, $D$ and $C$ in Table~\ref{table:DSL} are defined as tokens in this paper. We incorporate two types of layer blending modes—`normal' and `multiply'—to enhance visual variety in the output (controlled by corresponding tokens). To further constrain the search space, we divide the positions on a canvas into an 
{$8\times8$}
grid. Additionally, instead of using separate tokens in color channels (RGBA), we build a color list for $C$ that contains 54 different colors to choose from. This allows a single token to correspond to a wide range of colors, akin to a palette commonly used in modern drawing software and traditional paintings.
In summary, we define our goal of 2D graphics program inference as follows: reconstructing an input image (at a resolution of $64\times64$) by executing an inferred graphics program.

\section{Methodology}
In this section, we begin with a detailed description of the model architecture in Section~\ref{subsec:archi} and introduce the tokenization strategy in Section~\ref{subsec:tokenization} that enables our language model to interpret graphics programs as a sequence of tokens. Subsequently, we will discuss the training process in Section~\ref{subsec:train}, followed by an explanation of the inference pipeline in Section~\ref{subsec:inference}.
\begin{figure*}[t]
\centering

\begin{minipage}[b]{0.58\linewidth}
  \centering
 \centerline{\epsfig{figure=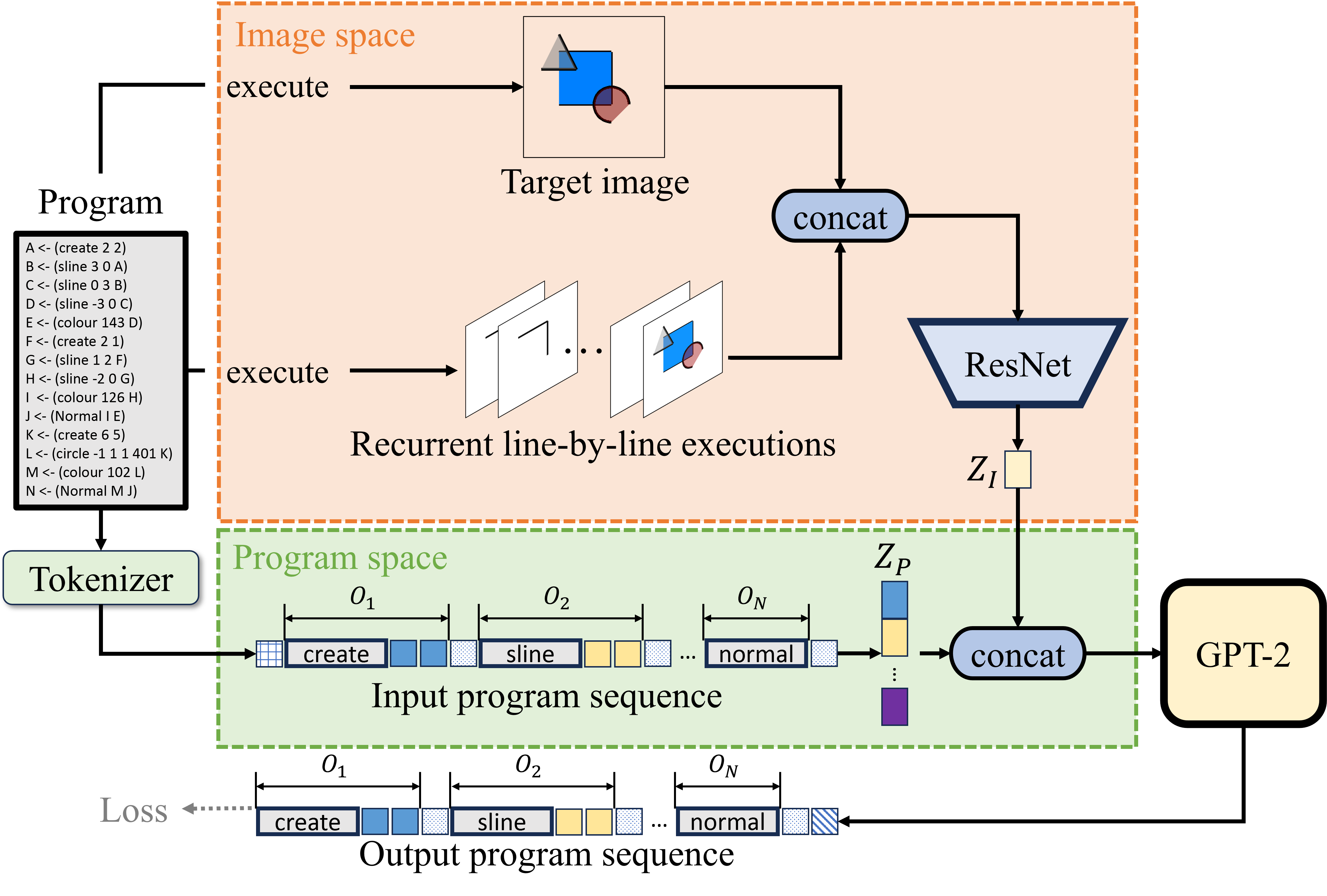,width=1\linewidth}}
  \centerline{(a)}\medskip
\end{minipage}
\begin{minipage}[b]{0.41\linewidth}
  \centering
  \centerline{\epsfig{figure=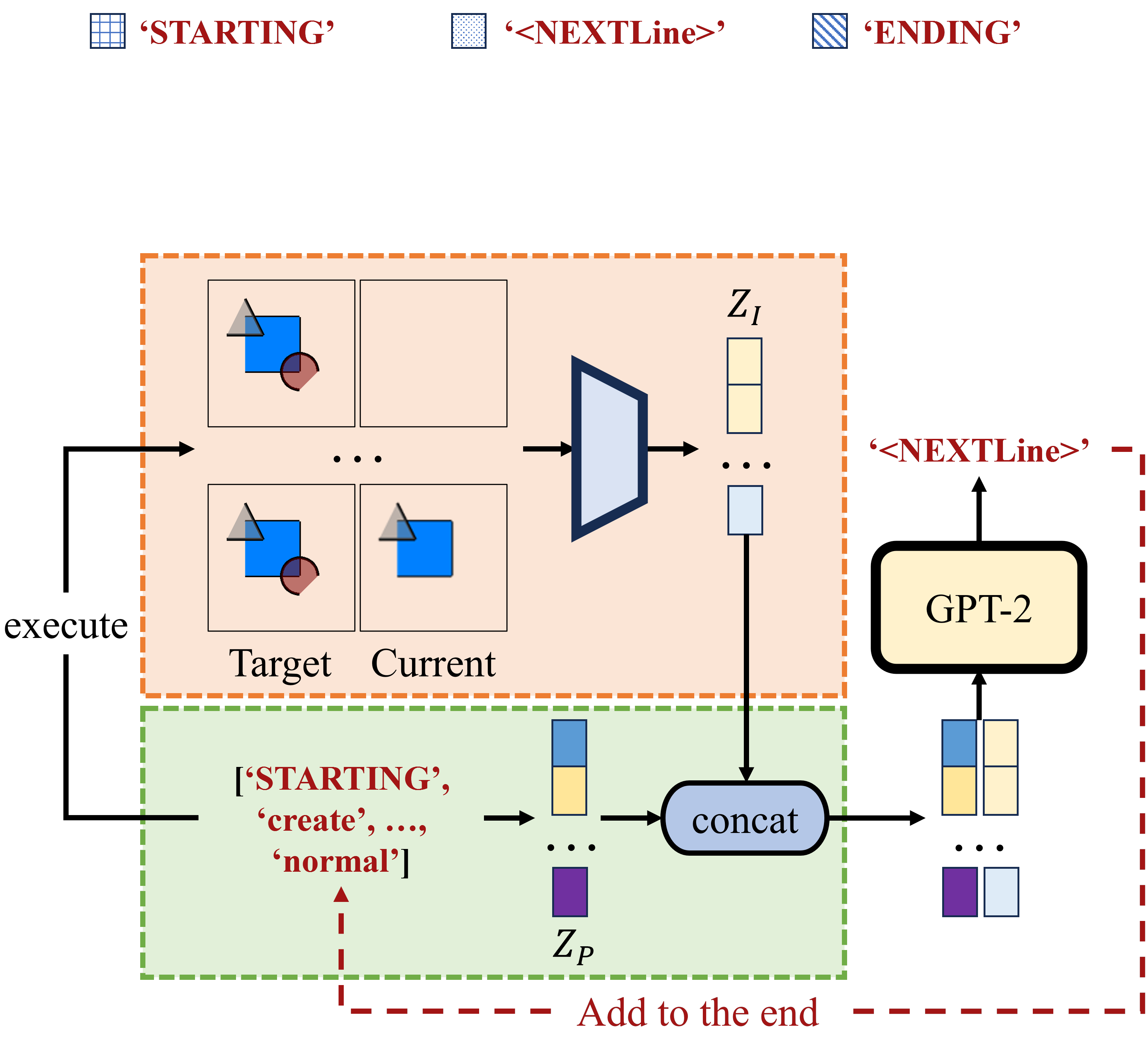,width=1\linewidth}}
  \centerline{(b)}\medskip
\end{minipage}

\caption{The overall architecture of Art2Prog, which contains two trainable modules: an ResNet-based image encoder and a program decoder built upon the architecture of GPT-2~\cite{gpt2}. (a): To reconstruct a target image into an executable graphics program, Art2Prog treats the program as a flattened sequence of tokens. It also utilizes image embeddings derived from the target image and partially executed programs as conditioning inputs. (b): Art2Prog infers a graphics program directly from the target image. By predicting the `ENDING' token, Art2Prog is capable of inferring programs of various lengths.
}
\label{fig:deep_architecture}
\end{figure*}
\subsection{Model architecture}
\label{subsec:archi}
We developed a deep architecture capable of efficiently inferring an executable graphics program from a randomly drawn image under our defined DSL. The comprehensive neural design of our network is illustrated in Fig.~\ref{fig:deep_architecture}(a), including two trainable modules: an image encoder (ResNet) to encode the executed image results and a program decoder (GPT-2) to learn the probability distribution of the tokenized program sequence. 
To bridge the gap between program syntax and semantics effectively, our approach integrates information from both image and program spaces. Inspired by REPL \cite{NEURIPS2019_50d2d226}, the graphics program is executed sequentially, line-by-line, to generate intermediate images through a non-differentiable, off-the-shelf rasterizer. These images are then concatenated with the target image, creating an $8$-channel input image to be fed into an image encoder. The encoder follows the architecture of ResNet without pretraining on other datasets. 

To process the graphics program, we flatten it into a sequence of tokens, appending a special token `$<$NEXTLine$>$' at the end of each line to signify its termination. This tokenized program sequence is then embedded and concatenated with the image embeddings on a token-wise basis, as illustrated in Fig.~\ref{fig:deep_architecture}(b). 
Given that each line of code can only be executed following the prediction of a `$<$NEXTLine$>$' token, each token within the same line of code is associated with the same executed image. Therefore, to enable the concatenation of images and programs in a practical manner, we duplicate the images to match the token length for each executed line of code. Only when a `$<$NEXTLine$>$' token is predicted is the execution result of the current program updated. 
We build our program decoder on the basis of an existing language model (GPT-2~\cite{gpt2}) to decode the concatenated embeddings of the current state into the program sequence for the next state. Additionally, we train our decoder from scratch without pretraining on other datasets. Consequently, we simplify the program inference problem by predicting the next token based on the current program sequence. The prediction distribution for a graphics program can thus be factorized as follows:
\begin{eqnarray}
p(S|I_T) &=& \prod \limits_{k=1}^K p(t_k|g_\theta \{ f_{\theta}(I_T, I_j), t_j \}_{j=1}^{k-1}) {,}
\end{eqnarray}
where $t_1...t_K$ are the tokens in the target flattened program sequence $S$, $K$ is the length of program sequence that varies across different programs, $g_\theta$ is the sequence decoder (GPT-2), and $f_\theta$ is the image feature extractor (we use ResNet-18 in this paper). $I_T$ and $I_j$ are the target image and canvas rendering at token $t_j$ respectively.

\subsection{Tokenization}
\label{subsec:tokenization}

Art2Prog employs a unique tokenization strategy to convert a graphics program into a sequence of tokens, facilitating feature concatenation and sequence decoding. The graphics program $P$ can be represented as $P = (O_1, ..., O_N)$, where $O_i$ indicates the $i^{th}$ line of code in the program. As mentioned in Table~\ref{table:DSL}, we first quantize arguments $X_i \in \{N, R, D, C\}$ into distinct intervals as tokens, and similarly, assign tokens to command classes $C_i$ (such as $create$, $sline$, $circle$, $color$, $normal$, and $mul$). Therefore, a single line of code $O_i$ may contain $2$ types of tokens: $O_i=(C_i, v_i^1, ..., v_i^k)$. Here, we do not tokenize the pointer to the former line $O_{i-1}$ because we execute our code line-by-line by default so that it can be simplified. For the layer combination commands $normal$ and $mul$, our detokenizer refers to $O_{i-1}$ and the second last $color$ command, which should indicate the end of a layer. Additionally, we introduce $3$ special tokens $\{ $`STARTING', `ENDING', `$<$NEXTLine$>$'$\}$ that indicate the start of a sequence, the end of a sequence, and the end of a line respectively.

\subsection{Training}
\label{subsec:train}
The primary training objective of our model is to minimize the cross-entropy loss for the predicted tokens at each position in the program sequence. Given the target program sequence $S$ and a corresponding target image $I_T$, we train our model $\Theta$ to minimize:
\begin{eqnarray}
l(\hat{S},S) &=& CE(\hat{S}, S | I_T; \Theta){,}
\end{eqnarray}
where $CE()$ refers to the cross-entropy function, and $\hat{S}$ is the output program sequence of our model.

Inspired by recent transformer-based language models, we shift the input sequence $s$ to the right by one position, as shown in Fig.~\ref{fig:deep_architecture}(a). Thus, the input sequence starts with a special token `STARTING' and the output sequence ends with a special token `ENDING'. For each line of code $O_n$, we assume that the preceding lines $(O_1, ..., O_{n-1})$ have been correctly successfully inferred; thus, we execute and render these partial lines to produce $n-1$ intermediate images $(I_1, ..., I_{n-1})$. As illustrated in Fig.~\ref{fig:deep_architecture}(b), these intermediate images, concatenated with the target image, are fed to ResNet to generate $n-1$ image embeddings. Since the lines are flattened to sequence before being fed into the GPT-2, the number of image embeddings should match the token length. Thus, we repeat each image embedding $I_i$ to the length $L_O+1$, where $L_O$ is the token length of the code to which these embeddings pertain. The addition of $1$ accounts for the additional special token `$<$NEXTLine$>$'. We then concatenate the extended image embeddings and program embedding in a token-wise manner prior to being fed into GPT-2.

\begin{algorithm}[t]
\caption{The inference process of Art2Prog.}
\label{alg:art2prog_inference}
\begin{algorithmic}
\REQUIRE ~~\\
the target RGBA image (spec)\\
\ENSURE ~~\\
\STATE Programs $P_{best}$
\STATE \textbf{Initialisation}: Start an empty program $P$. \\
\qquad \qquad \qquad \quad Start an empty sequence of tokens $O$ indicating current line of code. \\
\qquad \qquad \qquad \quad Set the maximum number of token in each line as $N_t$. \\
\qquad \qquad \qquad \quad Set the maximum number of lines in $P$ as $N_O$.
    \STATE $P \Leftarrow$ `STARTING'
    \REPEAT
        \IF{len($O$) > $N_t$}
            \STATE Reset $O$
        \ENDIF
        \STATE Samples the next token $t$ from current $P$ and $O$ 
        \STATE $O \Leftarrow t$
        \IF{$t$ is `<NEXTLine'}
            \STATE $P \Leftarrow O$
            \STATE Reset $O$
        \ENDIF
        \IF{$P_{best}$}
            \IF{Loss($P$) $<$ Loss($P_{best}$)}
                \STATE $P_{best} = P$
            \ENDIF
        \ELSE
            \STATE $P_{best}$ = $P$
        \ENDIF
    \UNTIL{len($P$) > $N_O$ or $t$ is `ENDING'}
\\
\textbf{return} $P_{best}$
\end{algorithmic}
\end{algorithm}

\subsection{Inference}
\label{subsec:inference}
As shown in Fig.~\ref{fig:deep_architecture}(b), our model infers a single token at a time, beginning with the initial input sequence [`STARTING'] and terminating with the prediction of `ENDING'. This design allows our model to infer graphics programs of varying lengths based solely on a target image as input. Our approach does not rely on specific search algorithms, such as beam search or Sequential Monte Carlo (SMC), which can significantly slow down the inference process. Instead, we employ a simple greedy search strategy while supporting early stopping if the program has already been correctly inferred (indicated by $MSE=0$). 

The implementation details of our graphics program inference scheme are shown in Algorithm 1. For each inference, we repeatedly conduct inference from empty until timeout. After that, we compare all of the generated programs by their $IoU_{rgba}$ distance from the target raster image to find the best match. The metric $IoU_{rgba}$ is defined as a modified version from $IoU$, which aims to measure the similarity of RGBA graphics with objects of different sizes.

\begin{figure}[t]
\centering

\includegraphics[width=0.7\linewidth]{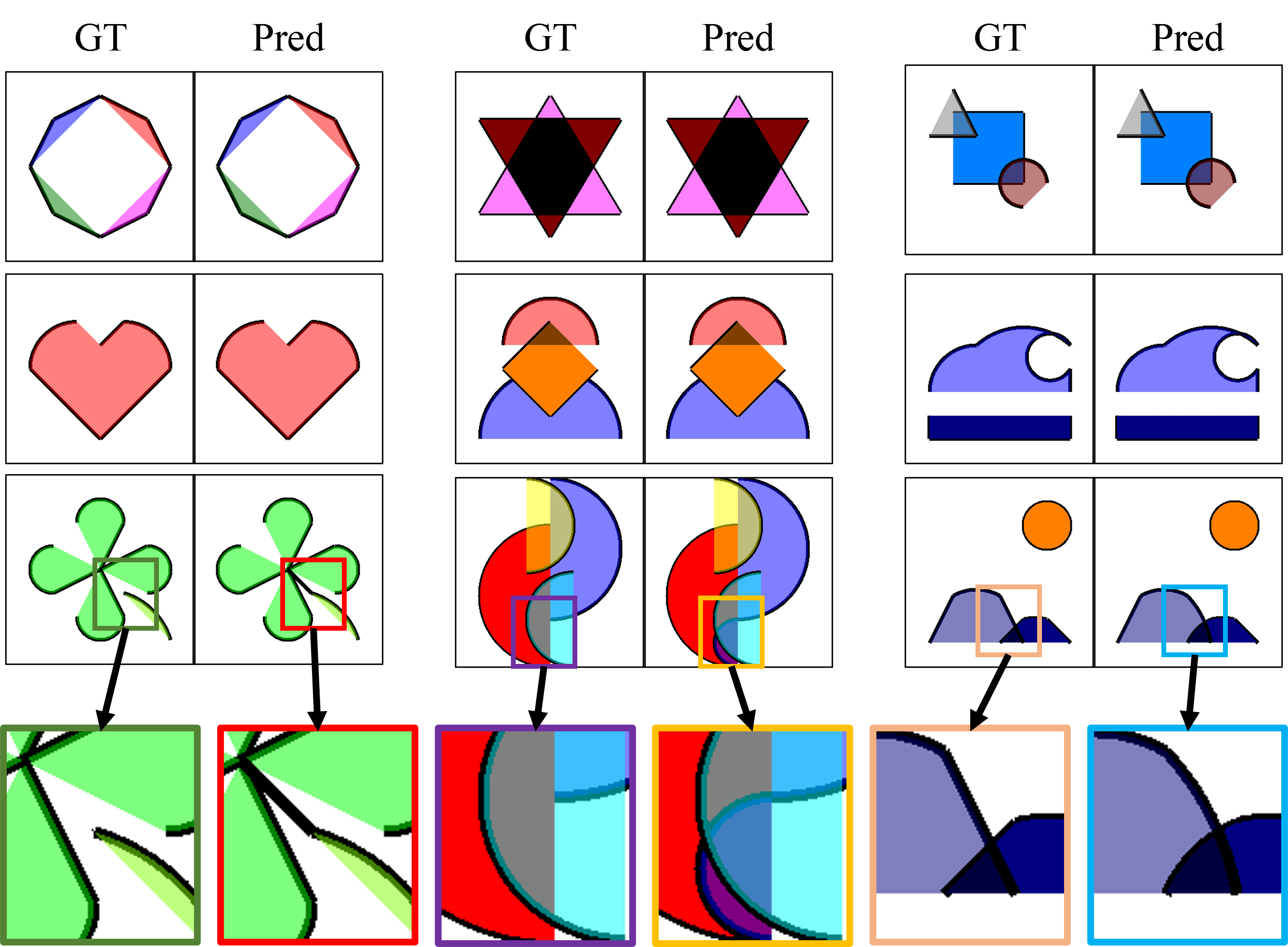}
\caption{Inference results on hand-drawn geometric artworks. We render all the images at resolution $300\times 300$ for better visualization. The input image size to the model is still $64 \times 64$. The bottom column shows examples of failure cases. } \label{fig:icons}
\end{figure}

Experimental results demonstrate that Art2Prog is capable of producing high-quality inferences within a short time frame (approximately $15$ seconds), surpassing current state-of-the-art techniques. We present examples of our inference results in Fig.~\ref{fig:icons}. The comparative analysis of results and a detailed implementation will be presented in the subsequent section.

\section{Experiments and Results}
\subsection{Data preparation and experiment settings}
We collect data for training by randomly generating graphics programs with up to 10 layers with reference to the defined DSL as described in Section~\ref{sec:DSL}. During training, our model was exposed to approximately 6 million examples. We utilize the Adam optimizer with a learning rate of $1\times10^{-3}$, using a batch size of $32$ across one RTX 3090 GPU for all settings in Table~\ref{tab:Ablation}. For evaluation, we built an eval set with 1000 generated data up to 13 layers in each program. Also, inspired by the design of $IoU$, we define a modified version to fairly compare the similarity among RGBA images with objects of different sizes:
\begin{equation}
  IoU_{rgba}(\hat{I}, I) = \frac{\sum_{p=1}^{P} (\hat{I}_{p}=I_{p})}{\sum_{p=1}^{P} (\hat{I}_{p}(A) > 0\ and\ I_{p}(A) > 0)},
  \label{equ:iourgba}
\end{equation}
where $\hat{I}$ and $I$ denote the predicted image and the target image, respectively. $\hat{I}_p$ and $I_p$ indicate the pixel value of $\hat{I}$ and $I$ at position $p$. $\hat{I}_p(A)$ and $I_p(A)$ are the pixel values at the alpha channel of images $\hat{I}$ and $I$, which indicate the transparency in the RGBA color space.

\begin{table*}[t]
\begin{center}
\caption{Quantitative results under different architecture designs of Arg2Prog. \#Layers denotes the maximum number of program layers each model was exposed to during training. SR denotes the rate of correct inferences ($IoU_{rgba}=1.0$). Max$_{PL}$ denotes the maximum program lengths. These scores are all derived from the same eval set, which comprises 1,000 data entries, each containing programs with up to 13 layers. We have equipped Monte Carlo Tree Search (MCTS) with models that utilize Pointer Network (PtrNet)~\cite{ptrnet} as the decoder. 
The inference timeout for all models is set at 125 seconds in this table.} 
\vspace*{1mm}
\label{tab:Ablation}
\centering
\begin{tabular}{c|ccc|ccccc}
  \hline
  Model & Encoder & Decoder & \#Layers (L)  & $IoU_{rgba}$ ($\uparrow$) & $IoU$ ($\uparrow$) & MSE ($\downarrow$) & SR ($\uparrow$) & Max$_\textit{PL}$ ($\uparrow$)\\
  \hline
  $A_{L=5}$ & CNNs & PtrNet & 5  & 0.7535 & 0.8471 & 0.1739 &0.593 & 26\\ 
  $B_{L=5}$ & RN&PtrNet & 5 & 0.9330 & 0.9692 & 0.0447 & 0.907 & 26\\
  $C_{L=5}$ & RN & GPT-2 & 5 & 0.9420 & 0.9858 & 0.0304 & 0.919 & 26\\ 
  \hline
  $A_{L=10}$ & CNNs&PtrNet & 10 & 0.5529 & 0.6722 & 0.2968 & 0.275 & 17\\ 
  $B_{L=10}$ & RN&PtrNet & 10& 0.9363 & 0.9719 & 0.0399 & 0.920 & 26\\
  $C_{L=10}$ &RN&GPT-2 & 10 & 0.9631 & 0.9923 & 0.0182 & 0.951 & 27\\
  \hline
\end{tabular}
\end{center}
\end{table*}

\begin{figure*}[t]
\centering

\includegraphics[width=0.9\linewidth]{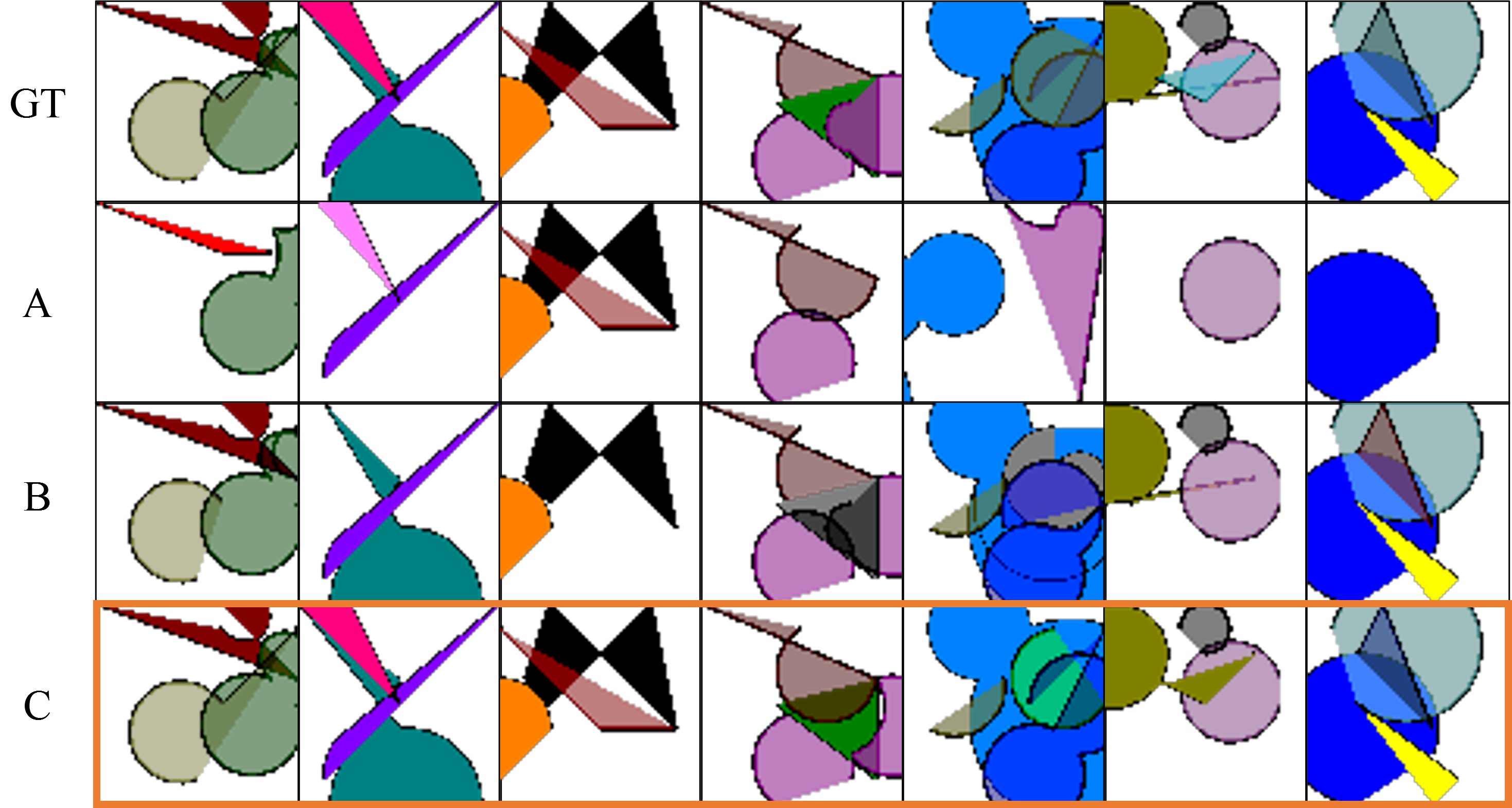}
\caption{Qualitative results of different architecture designs of Art2Prog. We pick the best models ($A_{L=5}$, $B_{L=10}$, and $C_{L=10}$) based on their average inference accuracy for comparison.} \label{fig:ablation}
\end{figure*}

\subsection{Ablations}

In order to ablate our architecture, we assessed the impacts of two key components in Art2Prog: the image encoder and the program generator. Additionally, we evaluated the influence of the training set on model performance. Our models are tested on an evaluation set comprising up to $13$ layers. Thus, we conducted training on two different sets, one with layers up to $5$ and another up to $10$, to determine whether training with longer programs enhances the performance of the synthesizer. 

We first compared the evaluation scores of different image encoders, including sequentially stacked Conv2d blocks with ReLU activation (CNNs) and ResNet-18. Notably, we introduced a minor modification to the original ResNet structure by removing Batch Normalization. This alteration led to a substantial improvement in performance. As indicated in the $A_{L=5}$ and $B_{L=5}$ of Table~\ref{tab:Ablation}, compared with CNNs, ResNet demonstrates higher reconstruction accuracy ($IoU_{rgba}$, $IoU$ and MSE) and the ability to precisely infer longer programs (with the highest rate of successfully inferred programs and maximum program length). When trained with longer programs (up to 10 layers), ResNet demonstrates a slight improvement in performance. Conversely, CNNs encounter difficulties in training under these conditions. By comparing the qualitative results in Fig.~\ref{fig:ablation}, it is evident that utilizing ResNet facilitates the construction of complex programs in the majority of scenarios. However, in specific instances, such as those illustrated in the third column of this figure, CNNs demonstrate superior performance in accurately reproducing a given image.

Subsequently, we demonstrated the necessity of using GPT-2 as the program generator, as opposed to Pointer Network (PtrNet)~\cite{ptrnet}. When utilizing PtrNet as the generator, the program is not flattened but treated as separate lines of code as in \cite{NEURIPS2019_50d2d226}.  We observed that GPT-2 consistently demonstrates superior accuracy, regardless of whether it is trained with longer program or not. Upon comparing the qualitative results depicted in the figure, it is observed that GPT-2 exhibits a marginally superior capability in preserving the sharp details within raster images. However, PtrNet achieves satisfactory performance in the majority of cases.
Overall, models trained with longer programs exhibit better generalization capabilities in complex scenes. 

\begin{figure}[t]
  \centering
\begin{minipage}{0.85\linewidth}
  \centering
 \centerline{\epsfig{figure=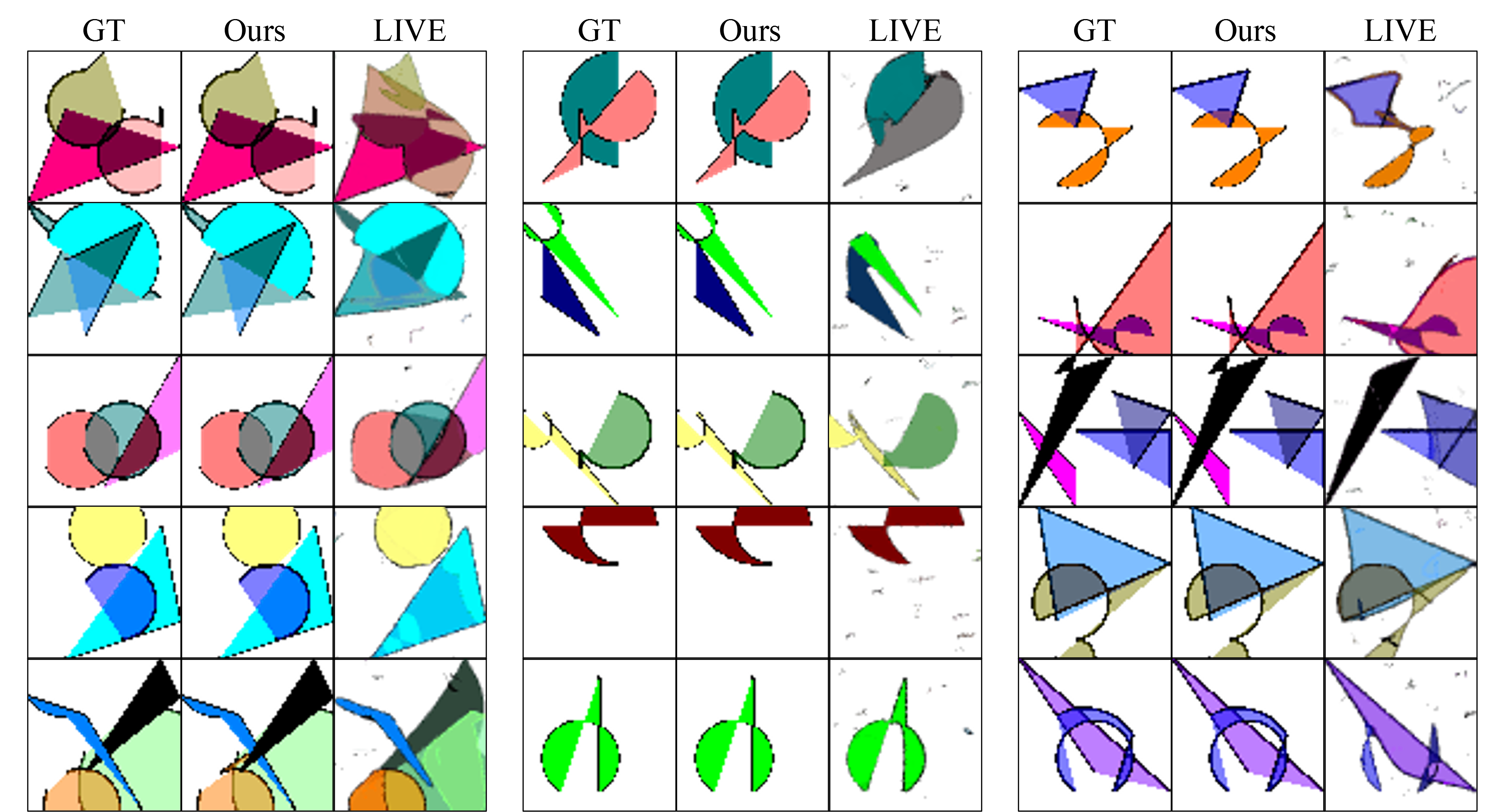,width=1\linewidth}}
  \centerline{(a)}\medskip
\end{minipage}
\begin{minipage}{0.85\linewidth}
  \centering
  \centerline{\epsfig{figure=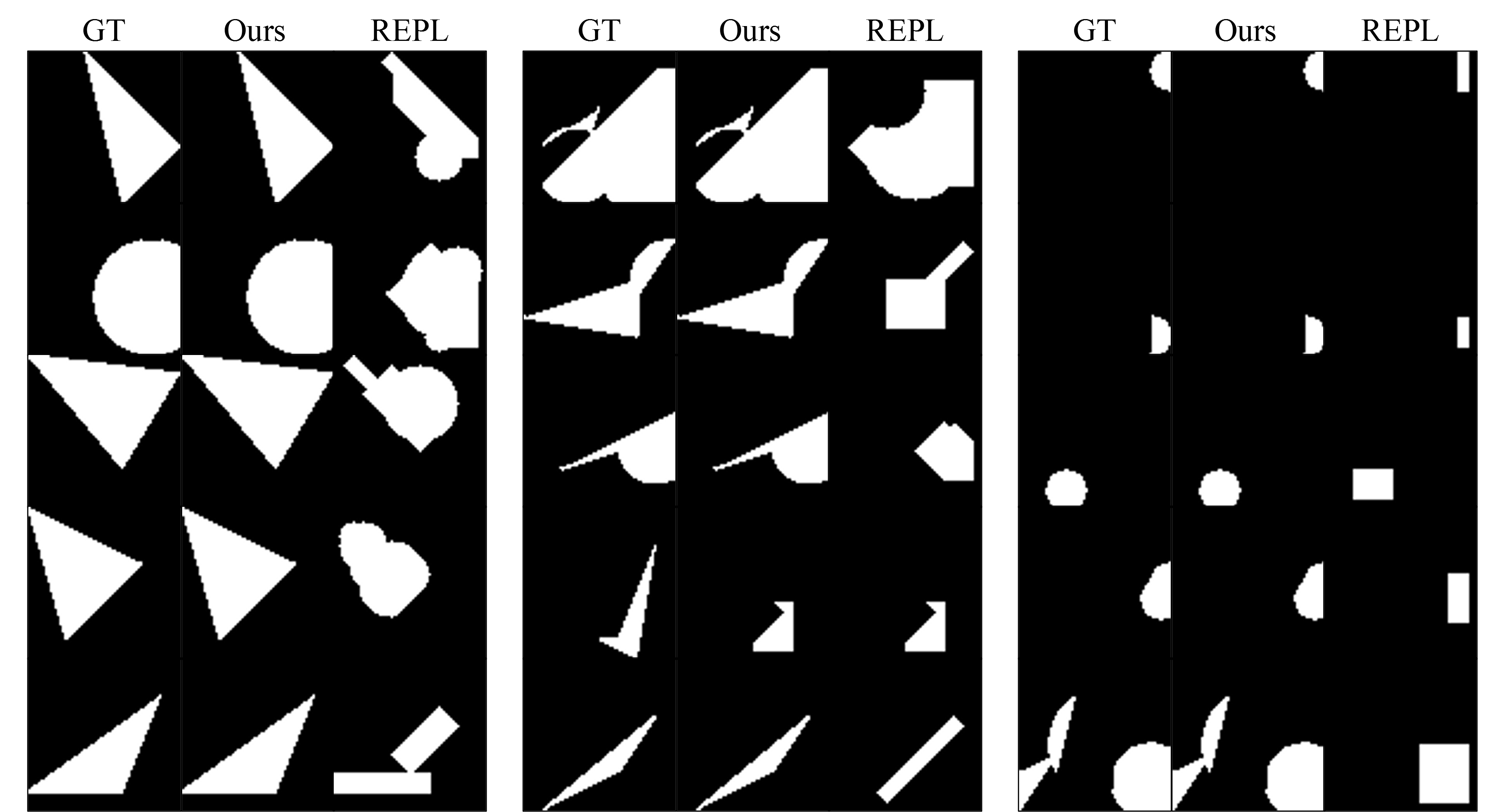,width=1\linewidth}}
  \centerline{(b)}\medskip
\end{minipage}
\caption{Our qualitative comparison with SOTA methods LIVE and REPL.}
\label{fig:SOTA}
\end{figure}

\begin{table}[t]
\begin{center}
\caption{Inference comparison on the evaluation set. Here we only picked the transparency channel for $IoU$ calculation. The timeout for each inference is 15 seconds in this table.}
\vspace*{1mm}
\label{tab:SOTA}
\begin{tabular}{lccc}
  \hline
  \multirow{2}{*}{Model} & Search  & \multirow{2}{*}{$IoU$ ($\uparrow$)} &  \multirow{2}{*}{MSE ($\downarrow$)}   \\
\ & Algorithm&&
  \\
  \hline
  REPL~\cite{NEURIPS2019_50d2d226} & GS  & 0.4514 & -\\
  REPL~\cite{NEURIPS2019_50d2d226}  & Beam & 0.5338  & -\\
  REPL~\cite{NEURIPS2019_50d2d226}  & SMC & 0.6630 & -\\
  \hline
  LIVE~\cite{LIVE} & - & 0.8513  & 0.0508\\
  Arg2Prog (ours) & GS & \textbf{0.9733}  & \textbf{0.0491}\\
  \hline
\end{tabular}
\end{center}
\end{table}

\subsection{Comparison with the state-of-the-art methods}
We compared our proposed method against two other state-of-the-art (SOTA) image vectorization methods: LIVE~\cite{LIVE} and REPL~\cite{NEURIPS2019_50d2d226}. We conducted this evaluation using a consistent dataset comprising 1000 data points, each generated under our DSL and containing up to 13 object layers per program. The results of this comparison are presented in Table~\ref{tab:SOTA} and Fig.~\ref{fig:SOTA}. 

Regarding the colored image-to-SVG method, LIVE exhibits inaccuracies in color inference, which negatively impacts its average pixel accuracy (MSE). Moreover, LIVE demonstrates difficulties in handling overlapped areas and in reconstructing paths that self-intersect. As illustrated in Fig.~\ref{fig:SOTA}(a), Art2Prog consistently maintains layer integrity in scenarios with overlapping layers and accurately reconstructs self-intersections. This figure showcases comparisons in various cases. The left column presents scenarios where LIVE overlooks details in overlapped layers of shapes, attributed to a misinterpretation of the segmentation process. The middle column reveals that LIVE performs poorly with areas of
self-intersection and often fails to capture sharp shapes accurately. The column on the right demonstrates cases involving both self-intersection and layer overlap, areas where LIVE struggles to achieve accurate reconstruction. In contrast, our approach is capable of comprehensively representing these complexities as graphics programs.

Since the REPL inverse CAD model supports only binary images as input, we processed the input data accordingly. As illustrated in Fig.~\ref{fig:SOTA}(b), it is evident that REPL struggles with the vectorization of complex and sharp shapes, often inaccurately interpreting shapes at the canvas's edges. In the examples showcased in the left column of this figure, REPL frequently generates redundant shapes to represent simple structures, such as a single triangle or a circle. For sharp shapes, as depicted in the middle column, REPL similarly faces difficulties in achieving accurate reconstruction. The right column demonstrates scenarios where, when tasked with predicting the graphics program for a shape positioned in a corner, REPL often inaccurately predicts a rectangle instead of the correct shape, such as a circle. In contrast, Art2Prog demonstrates precise program inference regardless of the shape's position and size.  

In summary, our model demonstrates superiority in color inference and the intricate reconstruction of compositions involving multiple shapes, producing outputs that are both clean and precise.

\section{Discussion}
\begin{figure*}[t]
\centering

\includegraphics[width=1\linewidth]{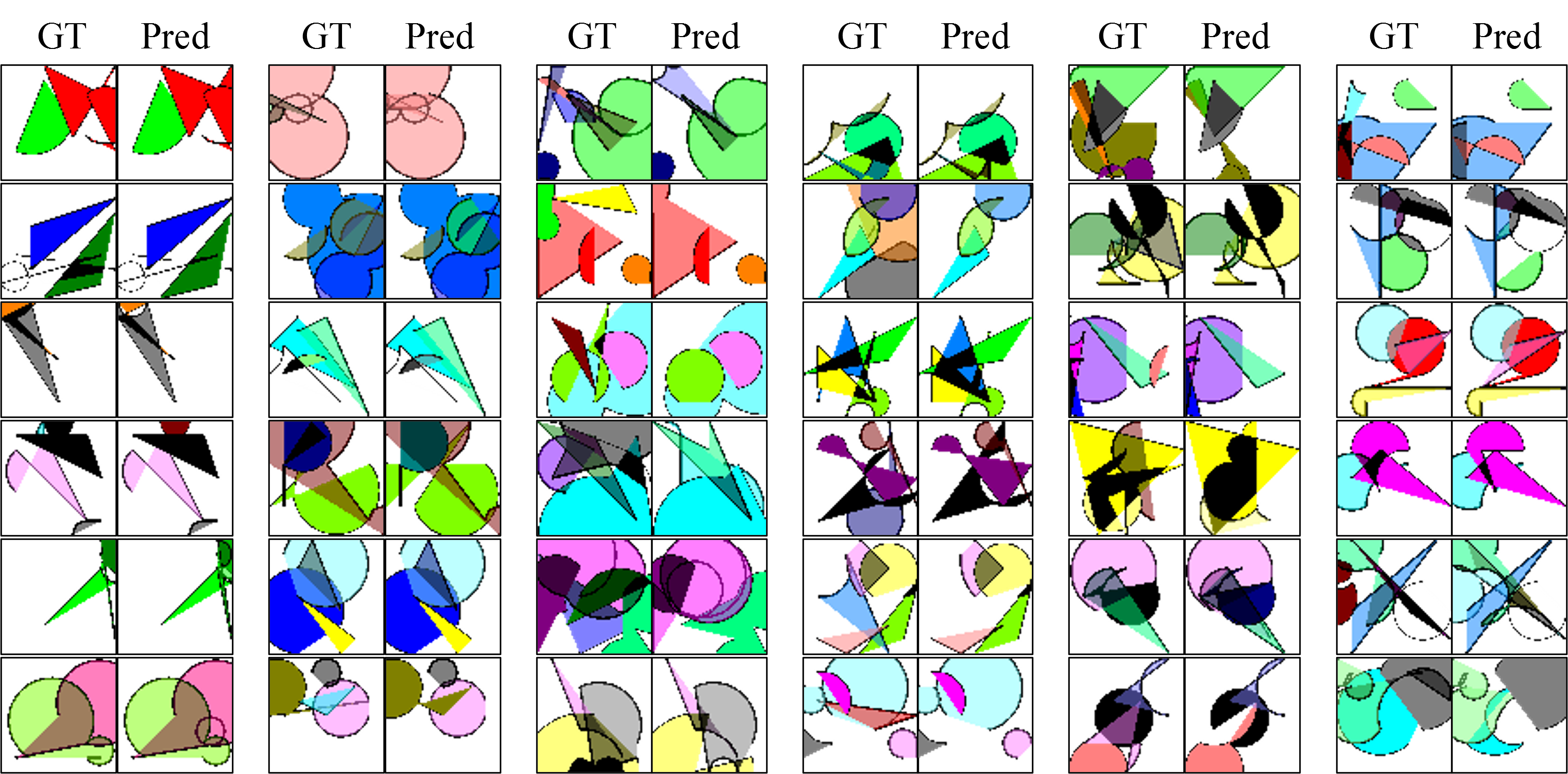}
\caption{Examples of failed cases, where the predicted graphics programs do not reproduce the target image comprehensively.} \label{fig:errors}
\end{figure*}

In this section, we examine the instances where our method failed to produce accurate predictions, as illustrated in Fig.~\ref{fig:errors}. These examples demonstrate failures in replicating the input raster image through generated graphic programs, particularly highlighting our system's tendency to neglect minor details, as evident  in the first column of examples in Fig.~\ref{fig:errors}. To improve accuracy, enhancing the image encoding component to more effectively capture detailed image features may be beneficial.

Failures in accurately determining the appropriate color blending mode or the exact color in scenarios where one layer completely encompasses another are depicted in the second column. These challenges stem from the ambiguity in distinguishing between blended colors and the dominant color of the upper layer. Incorporating a dedicated, trainable module specifically designed for layer combination might mitigate this issue. 

Moreover, in complex multi-layered images, our synthesis algorithm often misses segments, indicating difficulty in generating longer programs. This limitation points to potential advancements in program generation capabilities, possibly by exploring innovative architectural solutions or integrating more sophisticated large language models for future enhancement.

\section{Conclusion}
This paper presents a learning-based program synthesizer that aims to write a graphics program to represent the input image comprehensively. We propose a novel graphics program definition by separating the drawing steps towards a target RGBA image into several steps: (1) creating a new layer on the canvas, (2) outlining an enclosed shape with continuous lines, (3) filling the lined area in the current layer with colors and (4) combining layers with blending modes, which mimics the workflow of real-world digital artists. The quantitative and qualitative results in our experiments demonstrate that our approach achieves high-quality reconstruction results and effectively discerns the underlying drawing process. Though the drawing commands are highly simplified compared to existing drawing software, this paper can be considered as an initial step toward complex graphics program synthesis. Thus, future works can extend it to a broader range of more complex graphics.

\bibliographystyle{splncs04}
\bibliography{mybibliography}
\end{document}